\documentclass[sigconf]{acmart}

\usepackage{graphicx}
\usepackage{multirow} 
\usepackage{mathrsfs}
\usepackage{amsmath}
\usepackage{bm}     
\usepackage{makecell}
\usepackage{color}
\usepackage{amsfonts} 
\usepackage{float}
\usepackage{subfig}
\usepackage{balance}
\AtBeginDocument{%
  \providecommand\BibTeX{{%
    \normalfont B\kern-0.5em{\scshape i\kern-0.25em b}\kern-0.8em\TeX}}}

\copyrightyear{2022}
\acmYear{2022}
\setcopyright{acmcopyright}
\acmConference[MM '22] {Proceedings of the 30th ACM International Conference on Multimedia}{October 10--14, 2022}{Lisboa, Portugal.}
\acmBooktitle{Proceedings of the 30th ACM International Conference on Multimedia (MM '22), Oct. 10--14, 2022, Lisboa, Portugal}
\acmPrice{15.00}
\acmISBN{978-1-4503-9203-7/22/10}
\acmDOI{10.1145/3503161.3547974}

\begin{document}


\title{Unified Multimodal Model with Unlikelihood Training for\\ Visual Dialog}



\author{Zihao Wang} 
\email{zhwang_tjuer@tongji.edu.cn} 
\author{Junli Wang} 
\email{junliwang@tongji.edu.cn}
\author{Changjun Jiang}
\authornote{Changjun Jiang is the corresponding author.}
\email{cjjiang@tongji.edu.cn}
\affiliation{%
  \institution{Key Laboratory of Embedded System and Service Computing (Tongji University), Ministry of Education}
  \institution{National (Province-Ministry Joint) Collaborative Innovation Center for Financial Network Security, Tongji University}
  \city{Shanghai}
  \country{China}
}




\renewcommand{\shortauthors}{Zihao Wang, Junli Wang, \& Changjun Jiang}
\renewcommand{\authors}{Zihao Wang, Junli Wang, Changjun Jiang}
\renewcommand{\shorttitle}{Unified Multimodal Model with Unlikelihood Training for Visual Dialog}

\begin{abstract}
  The task of visual dialog requires a multimodal chatbot to answer sequential questions from humans about image content. Prior work performs the standard likelihood training for answer generation on the positive instances (involving correct answers). However, the likelihood objective often leads to frequent and dull outputs and fails to exploit the useful knowledge from negative instances (involving incorrect answers). In this paper, we propose a Unified Multimodal Model with UnLikelihood Training, named UniMM-UL, to tackle this problem. First, to improve visual dialog understanding and generation by multi-task learning, our model extends ViLBERT from only supporting answer discrimination to holding both answer discrimination and answer generation seamlessly by different attention masks. Specifically, in order to make the original discriminative model compatible with answer generation, we design novel generative attention masks to implement the autoregressive Masked Language Modeling (autoregressive MLM) task. And to attenuate the adverse effects of the likelihood objective, we exploit unlikelihood training on negative instances to make the model less likely to generate incorrect answers. Then, to utilize dense annotations, we adopt different fine-tuning methods for both generating and discriminating answers, rather than just for discriminating answers as in the prior work. Finally, on the VisDial dataset, our model achieves the best generative results (69.23 NDCG score). And our model also yields comparable discriminative results with the state-of-the-art in both single-model and ensemble settings (75.92 and 76.17 NDCG scores).
\end{abstract}





\maketitle

\begin{figure}[t]
  \centering
  \includegraphics[width=0.90\linewidth]{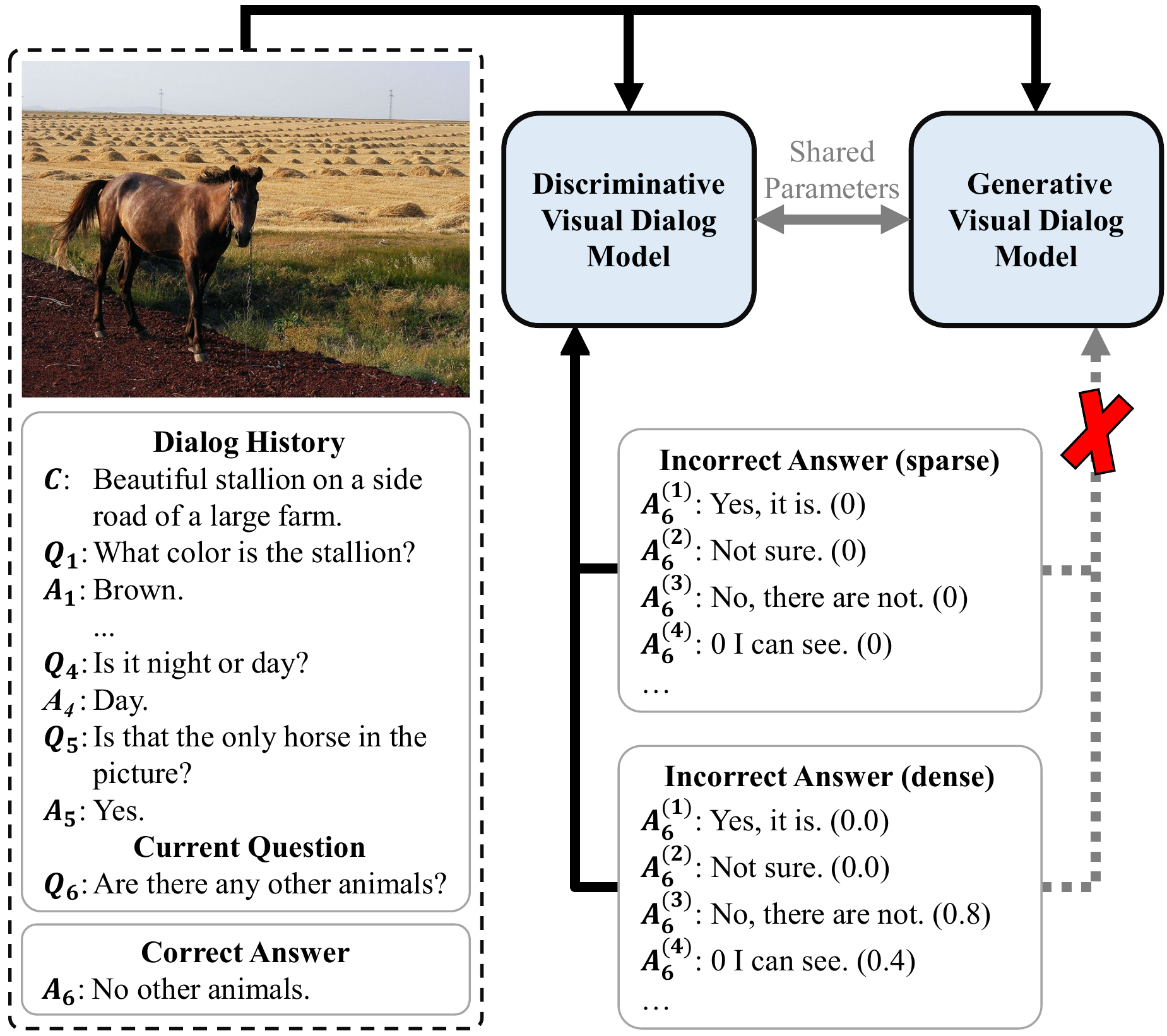}  
  \caption{\small
    Overview of the visual dialog task and existing models.
    (1) An example of the visual dialog task (left) is shown.
    (2) Existing visual dialog models (top right) include discriminative and generative models, where two models could share the parameters. 
    (3) Incorrect answers (bottom right) are only used to train the discriminative model, but the generative model fails to exploit incorrect answers (dashed arrow with red cross). 
    This paper considers using incorrect answers for training both discriminative and generative models.
 }  
  \label{fig1}
\end{figure}

\section{Introduction}

With the development of multimodal machine learning~\cite{MMML}, researchers have proposed vision-language tasks such as image captioning~\cite{Anderson_2018_ImageCaptioning}, visual question-answering~\cite{Antol_2015_VQA}, and visual dialog~\cite{VisualDialog_2017_Das} to drive the merging of vision and language. 
Specifically, visual dialog~\cite{VisualDialog_2017_Das} is proposed to simulate multi-round human-chatbot interaction using natural language in the multimodal environment. 
This task requires the chatbot to answer the current question based on an image and a dialog history that involves caption and prior question-answer pairs. 
Considering the dialog example in Figure~\ref{fig1} (left), given the image, the dialog history and the current question $Q_6$, the visual dialog model needs to choose or generate an appropriate answer $A_6$.
Besides, there are two types of annotations for the candidate answer: sparse annotation is 1 or 0, indicating whether it is the original correct answer; dense annotation ranges from 0.0 to 1.0, indicating relevance score. 
The visual dialog models have rich application scenarios, such as helping the visually impaired, interacting with intelligent assistants, operating robots, and information retrieval from large-scale visual media data via language.

As shown in Figure~\ref{fig1} (top right), there are two settings for visual dialog models: discriminative and generative. In the discriminative setting, the model performs the answer discrimination task to give a ranking of all candidate answers, and in the generative setting, the model performs the answer generation task.  
Besides, researchers have observed that multi-task learning effectively improves model performance, which shares the parameters between generative and discriminative models and trains the entire model using both tasks.
Specifically,
some researchers~\cite{nguyen2020LTMI, gog_2021_chen} train the entire model with two decoders simultaneously. And Wang et al.~\cite{VDBERT_2020_wang} propose a unified model that supports both discriminative and generative settings without explicit decoders. 
However, all of them use likelihood training on positive instances and fail to consider the incorrect answers in the generative setting. 
Since models trained by the likelihood objective tend to assign too much probability to frequent and dull tokens~\cite{welleck2019unlikelihood}, we expect to exploit useful knowledge from incorrect answers to attenuate this adverse effect.
Among these incorrect answers, as shown in Figure~\ref{fig1} (bottom right), there are wrong answers (e.g., "\emph{Yes, it is.}") that are the opposite of the correct answer (i.e., "\emph{No other animals.}"), frequent and dull answers (e.g., "\emph{Not sure.}"), and acceptable answers (e.g., "\emph{No, there are not.}") that are relevant to the correct answer. 
By utilizing both correct and incorrect answers, our goal is to make the model tend to generate acceptable answers and reduce the probability of synthesizing wrong and dull answers.
 
In this paper, under the framework of the ViLBERT model (a two-stream transformer), we propose a unified model that holds both answer discrimination and answer generation seamlessly thanks to two kinds of attention masks. Besides, the model incorporates unlikelihood training for improving visual dialog generation.

First, the way of obtaining a unified model is to optimize the model in generative and discriminative settings jointly. 
We use a two-stream multi-modal model, ViLBERT, as the base model. 
We expect both settings are supported in the same model with different attention masks and tasks, where the attention masks control each token's access to the context. 

Second, to implement the autoregressive MLM task for answer generation, we design novel generative attention masks.   
For an answer sequence, VD-BERT only randomly masks out and predicts 15\% tokens of the answer each time during training. 
By contrast, inspired by XLNET~\cite{Yang2019XLNet} and UniLMv2~\cite{Bao2020unilmv2}, to perform the autoregressive MLM task to predict all the tokens of the answer, we append additional mask sequence, which contains multiple special mask tokens \texttt{[M]} and has the same length as the answer, after the dialog sequence.  
Moreover, to make the model less likely to synthesize incorrect answers, we utilize unlikelihood training, which is combined with likelihood training as the objective for autoregressive MLM, to take advantage of negative instances. 


Finally, to obtain a fine-tuned unified model using dense annotations, we propose a generative fine-tuning method for answer generation and use a ranking module as the fine-tuning method for answer discrimination.
As shown in Figure ~\ref{fig1}, among these incorrect candidate answers with dense annotations, there are some answers (e.g., "\emph{No, there are not.}"
) that are highly relevant to the correct answer.
Existing fine-tuning methods~\cite{VDBERT_2020_wang, VisDialBert_2020_Vishvak} only use dense annotations in discriminative setting. Since likelihood training fails to use all dense annotations directly, we propose a new fine-tuning objective for answer generation, which extends the objective of autoregressive MLM by considering candidates' relevance scores.

In summary, our contributions are listed as follows:
\begin{itemize}
\item To take advantage of multi-task learning, we propose a two-stream unified multimodal model that supports both generative and discriminative settings. We show the case that multi-task learning improves the performance of the visual dialog model. 

\item 
To reduce the probability of generating incorrect answers, our model incorporates unlikelihood training via the autoregressive MLM task. And to realize this task in the generative setting, we propose generative attention masks.

\item 
To exploit dense annotations in both settings, we not only select a state-of-the-art discriminative fine-tuning method but also propose a generative fine-tuning method.

\item Our model achieves new state-of-the-art generative results and comparable discriminative results on VisDial benchmarks. Besides, we perform the ablation study and case study to analyze the influences of the proposed mechanisms.

\end{itemize}

\section{Related Work}
 

\subsection{Visual dialog} 
\label{re_1}
The visual dialog task is first proposed by Das et al.~\cite{VisualDialog_2017_Das}, and early visual dialog methods obtain the features of images and texts via CNN and LSTM. The features are then merged by kinds of mechanisms such as using multi-modality graph~\cite{cheng2022financial} to bridge the cross-modal semantic relations between vision and text knowledge\cite{Zheng_2019_CVPR, jiang2020kbgn, gog_2021_chen, jiang2020dualvd, Exploring_2021_Li}, designing efficient attention mechanisms~\cite{gan2019multistep, nguyen2020LTMI, schwartz2019fga, Niu_2019_CVPR}, and incorporating causal principles~\cite{Qi_2020_P1P2}, to perform multimodal answer reasoning or generation. 
Since these models are trained individually on the VisDial dataset without using other large-scale datasets, they tend to perform less well than the pre-trained models.
 
Following the significant development of large-scale pre-trained models~\cite{Devlin2019BERT} in the field of natural language processing, researchers have designed multimodal transformer-based models~\cite{lu2019vilbert, tan2019lxmert, li2020unicodervl}.
Murahari et al.~\cite{VisDialBert_2020_Vishvak} adapt the two-stream ViLBERT~\cite{lu2019vilbert} to visual dialog via a two-step training and provide a simple state-of-the-art baseline VisDial-BERT. However, VisDial-BERT only supports answer discrimination. 
Wang et al.~\cite{VDBERT_2020_wang} propose a single-stream unified VD-BERT that seamlessly supports both discriminative and generative settings through three-step training.
However, during training, VD-BERT only learns 15\% tokens of the answer each time and only holds one task (masked language modeling) in the generative setting. 
Recently, \cite{VDPCR_2022_Yu} proposes VD-PCR, which uses VisDial-BERT as the base model and achieves state-of-the-art discriminative results by exploiting the Pronoun Coreference Resolution (PCR) task~\cite{Kottur_2018_ECCV}, and yet VD-PCR requires additional PCR labels and also fails to support answer generation.
Furthermore, to take advantage of dense annotations, existing work uses the fine-tuning methods only for answer discrimination; we adopt different fine-tuning methods for both answer discrimination and answer generation.  

\subsection{Unlikelihood Training}
\label{re_2}
Welleck et al.~\cite{welleck2019unlikelihood} point out that standard likelihood objective results in dull and repetitive generations and first propose token and sequence level unlikelihood training for text generation. 
In the area of generative dialog, some studies have focused on improving the standard likelihood training approach. 
Li et al. \cite{li2020dont} show applying unlikelihood to collected data of what a model should not do is effective for improving logical consistency. 
He and Glass~\cite{glass2020negativeTraining} propose a framework named negative training to minimize undesirable behaviors, such as malicious responses and generic responses.  
Inspired by the idea of~\cite{li2020dont, glass2020negativeTraining}, we provide a new way of using sequence-level unlikelihood training~\cite{welleck2019unlikelihood} on the visual dialog model to reduce the probabilities of wrong answers.

\section{Preliminary} 
\subsection{The visual dialog task}

Given a current question $Q_t$ grounded on the context that involves an image $I$ and dialog history $H = \{C, (Q_1, A_1), \dots, (Q_{t-1}, A_{t-1})\}$, the visual dialog task is to provide a ranking of a list of $N_c$ candidate answers $\{A^{(k)} _t\}_{k=1}^{N_c}$. 
Each original visual dialog consists of an image $I$, an image caption $C$, and a series of question-answer pairs $\{(Q_{t}, A_{t})\}_{t=1}^{N_t}$ about the image. 
Besides, sparse annotations and dense annotations for candidate answers are provided.
Sparse annotations label the original correct answer $A_t$ among all candidate answers. Dense annotations label each candidate with a dense relevance score between 0.0 and 1.0, i.e., $s(A^{(k)}_t) \in [0,1]$.

In general, there are two types of models to predict the answer: discriminative models that only score the given candidates, e.g. via a softmax over all candidates, or generative models that generate an answer, e.g. via a language model, and the log-likelihood scores are used to sort the candidate answers. For details, please refer to~\cite{VisualDialog_2017_Das}.

\subsection{Input Representation}
We concatenate the image token sequence from $I$ and the dialog sequence $(C, Q_1, A_1, \cdots, Q_t, \hat{A}_{t}) $ together as input. The image token sequence is taken as visual input, and the dialog sequence is taken as linguistic input. Specifically, 
the dialog sequence contains history $H$=\{$C$, $Q_1$, $A_1$, $\cdots$, $Q_{t-1}$, $A_{t-1}$\}, the current question $Q_t$, and a candidate answer $\hat{A}_{t}$ sampled from all candidate answers $\{A_t^{(k)}\}_{k=1}^{N_c}$.
A special token \texttt{[CLS]} is added at the beginning of the dialog sequence, and each question and answer are separated by a special token \texttt{[SEP]}.
The CNN features of object bounding boxes~\cite{fasterrcnn} in the image are extracted as image tokens.

\subsection{Backbone: two-stream Transformer} 

We adopt ViLBERT~\cite{lu2019vilbert} as the base model, which extends BERT~\cite{Devlin2019BERT} to a two-stream transformer for jointly modeling linguistic and visual features.
The input linguistic vectors $\mathbf{x}=\{x_{i}\}_{i=1}^{T}$ and visual features $\mathbf{y}=\{y_{i}\}_{i=1}^{\mathcal{T}}$ are packed in to $\mathbf{H}^0_{X}$=$[x_1, \cdots, x_{T}] \in \mathbb{R}^{T \times d_{x}}$ and $\mathbf{H}^0_{Y}$=$[y_1, \cdots, y_{\mathcal{T}}]  \in \mathbb{R}^{\mathcal{T} \times  d_{y}}$. The model obtains final representations 
$\mathbf{H}_{X}$ and $\mathbf{H}_{Y}$ using the stacked two-steam transformer layers.
In each layer, there are self-attention or co-attention heads, followed by a feed-forward network.
\paragraph{{Self-Attention and Co-Attention Masks}} 
Take the linguistic stream as an example. The output $A_X^l \in \mathbb{R}^{T\times d_k}$ of a self-attention (a) or co-attention (b) head of the $l$-th transformer layer is:
\begin{equation} 
\label{eq1}
\small
\begin{split}
   \mathbf{Q}&=\mathbf{H}^{l-1}_X\mathbf{W}^Q_{l},\quad
   \mathbf{K}=\left\{
                \begin{aligned}
                    & \mathbf{H}^{l-1}_X \mathbf{W}^{XK}_{l},
                    \text{(a)}\\
                    & \mathbf{H}^{l-1}_Y \mathbf{W}^{YK}_{l} ,
                    \text{(b)}
                \end{aligned}
                \right., \quad
   \mathbf{V}=\left\{
                \begin{aligned}
                    & \mathbf{H}^{l-1}_X\mathbf{W}^{XV}_{l}, \text{(a)}\\
                    & \mathbf{H}^{l-1}_Y\mathbf{W}^{YV}_{l},
                    \text{(b)}
                \end{aligned}
                \right., 
                \\ 
   \mathbf{M}_{i,j} &= \left\{
                            \begin{aligned}
                                &0,  \qquad\text{allow to attend}   \\
                                &-\infty,   \ \ \text{otherwise} 
                            \end{aligned}
                        \right., 
   \quad \,\,\,\,                    
   \mathbf{A}_X^l = \text{softmax}(\frac{\mathbf{QK^\top}}{\sqrt{d_k}}+\mathbf{M})\mathbf{V},
\end{split}
\end{equation}
where $\mathbf{W}^Q_l\in \mathbb{R}^{d_{x}\times d_k}$, $\mathbf{W}^{XK}_l$, $\mathbf{W}^{XV}_l$, $\mathbf{W}^{YK}_l$, $\mathbf{W}^{YV}_l$ are parameter matrices. 
$\mathbf{H}^{l-1}_X$ and $\mathbf{H}^{l-1}_Y$ are the previous layer’s linguistic and visual outputs. 
For the self-attention (a) head, $\mathbf{H}^{l-1}_X$ is projected to \emph{keys} $\mathbf{K}$ and \emph{values} $\mathbf{V}$ using $\mathbf{W}^{XK}_l$, $\mathbf{W}^{XV}_l \in \mathbb{R}^{d_{x}\times d_k}$, and the mask matrix $\mathbf{M} \in \mathbb{R}^{T \times T}$ determines whether language tokens can attend to each other. 
For the co-attention (b) head, $\mathbf{H}^{l-1}_Y$ is projected to \emph{keys} $\mathbf{K}$ and \emph{values} $\mathbf{V}$ using $\mathbf{W}^{YK}_l$, $\mathbf{W}^{YV}_l \in \mathbb{R}^{d_{y}\times d_k}$, and $\mathbf{M} \in \mathbb{R}^{T \times \mathcal{T}}$ determines whether language tokens can attend to vision tokens.

\section{Method}

\begin{figure*}[t]
  \centering 
  \includegraphics[width=0.93\linewidth]{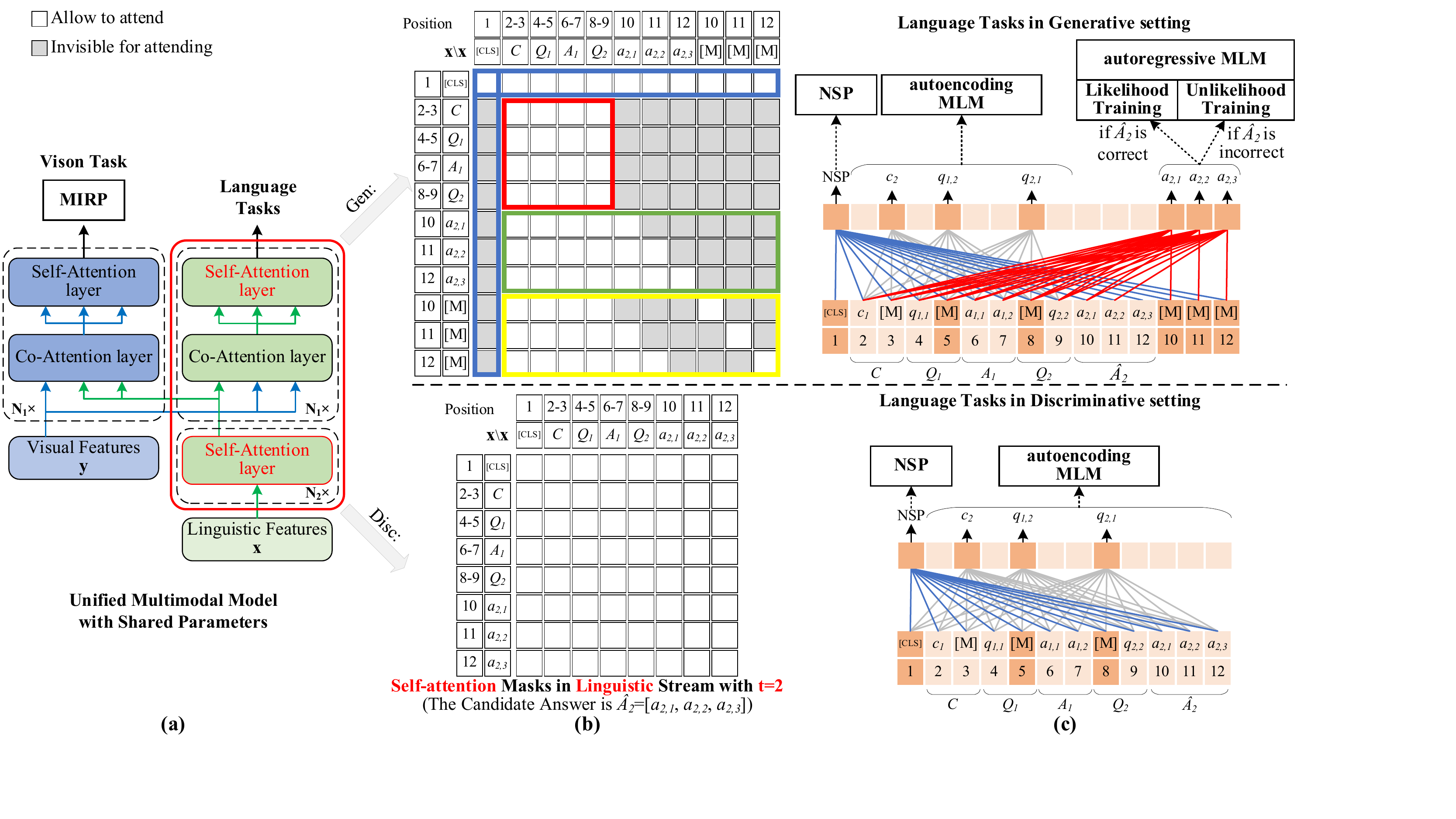}  
  \caption{ 
    Overview of UniMM-UL. (a) The model has both visual and linguistic streams for vision and language tasks. And model parameters are shared between generative (Gen) and discriminative (Disc) settings.
    (b) In the linguistic stream, we show the self-attention masks in two settings, which control what linguistic context a language token can access.
    (c) Various Language tasks are used in two settings, where NSP and autoencoding MLM are used in both settings, and autoregressive MLM is used only in the generative setting.
   } 
  \label{fig2}
\end{figure*}
    
In this section, we formally introduce our proposed Unified Multimodal Model with UnLikelihood training (UniMM-UL) that supports both generative and discriminative settings, and the overview is shown in Figure~\ref{fig2}.  
To make the model support the generative setting, we first propose the generative attention masks (Section \ref{section_generative_masks}).  
Then, we exploit several tasks (i.e., autoregressive MLM, autoencoding MLM, Next Sentence Prediction (NSP) and Masked Image Region Prediction (MIRP)) to train the unified model in both settings with different attention masks.
Especially, to utilize the negative instances, we incorporate unlikelihood training into the model by the autoregressive MLM task (Section \ref{Language_Tasks}). 
Finally, to conduct unified fine-tuning using dense annotations, we propose a generative fine-tuning method and adopt a state-of-the-art ranking module as a discriminative fine-tuning method (Section \ref{finetuning_dense}).

\subsection{Generative Attention Masks}
\label{section_generative_masks}
In order to implement autoregressive MLM, autoencoding MLM, and NSP tasks simultaneously in the generative setting, we propose generative masks for each attention in our two-stream model. As each stream contains two types of attention (i.e., co-attention and self-attention), there are four types of attention masks in the two-stream transformer. 
We first design the generative self-attention mask in the linguistic stream (Section \ref{section_generative_self_masks}). Then, we describe other three generative attention masks (Section~\ref{section_generative_other_masks}).

\subsubsection{Generative Self-Attention Mask in the Linguistic Stream}
\label{section_generative_self_masks}
To implement multiple tasks simultaneously in the generative setting, we design the generative self-attention mask.
We introduce the following four cases corresponding to the four boxes with colors in the mask, as shown at the top of Figure 2(b).

First, to get autoregressive attention, we make each token of $\hat{A}_t$ can not access the "future" tokens (green box) by setting $-\infty$ to the elements of $\mathbf{M}$ in Eq.~(\ref{eq1}). 
Next, to conduct autoencoding MLM on the past tokens (i.e., $C$, $Q_1$, $A_1$, \dots, $Q_t$), we allow these tokens to attend to each other (red box) by setting $0$ to the elements of $\mathbf{M}$.
Then, to learn all tokens of $\hat{A}_t$ each time during training, we append multiple \texttt{[M]} tokens with positions to the end of the dialog sequence, where each \texttt{[M]} attends to itself and prior language tokens (yellow box). 
And the final representations of these \texttt{[M]} tokens are used to predict tokens of $\hat{A}_t$ in the corresponding positions.
Finally, in order to realize NSP as an auxiliary task in the generative setting, the \texttt{[CLS]} token could attend to all other language tokens, but other language tokens cannot attend to it (blue boxes). Otherwise, the information of $\hat{A}_t$ will be revealed through \texttt{[CLS]} during the training. 
Differing from the Seq-to-Seq model~\cite{Li2020unilmv1, VDBERT_2020_wang} that learns 15\% answer's tokens each time and could not use the NSP task, our model learns all tokens of each answer each time and supports the NSP task.
Please refer to appendix~\ref{appendix} for a detailed comparison.

\subsubsection{Other Generative Attention Masks.}
\label{section_generative_other_masks}
\begin{figure}[t]
  \centering
  \includegraphics[width=0.56\linewidth]{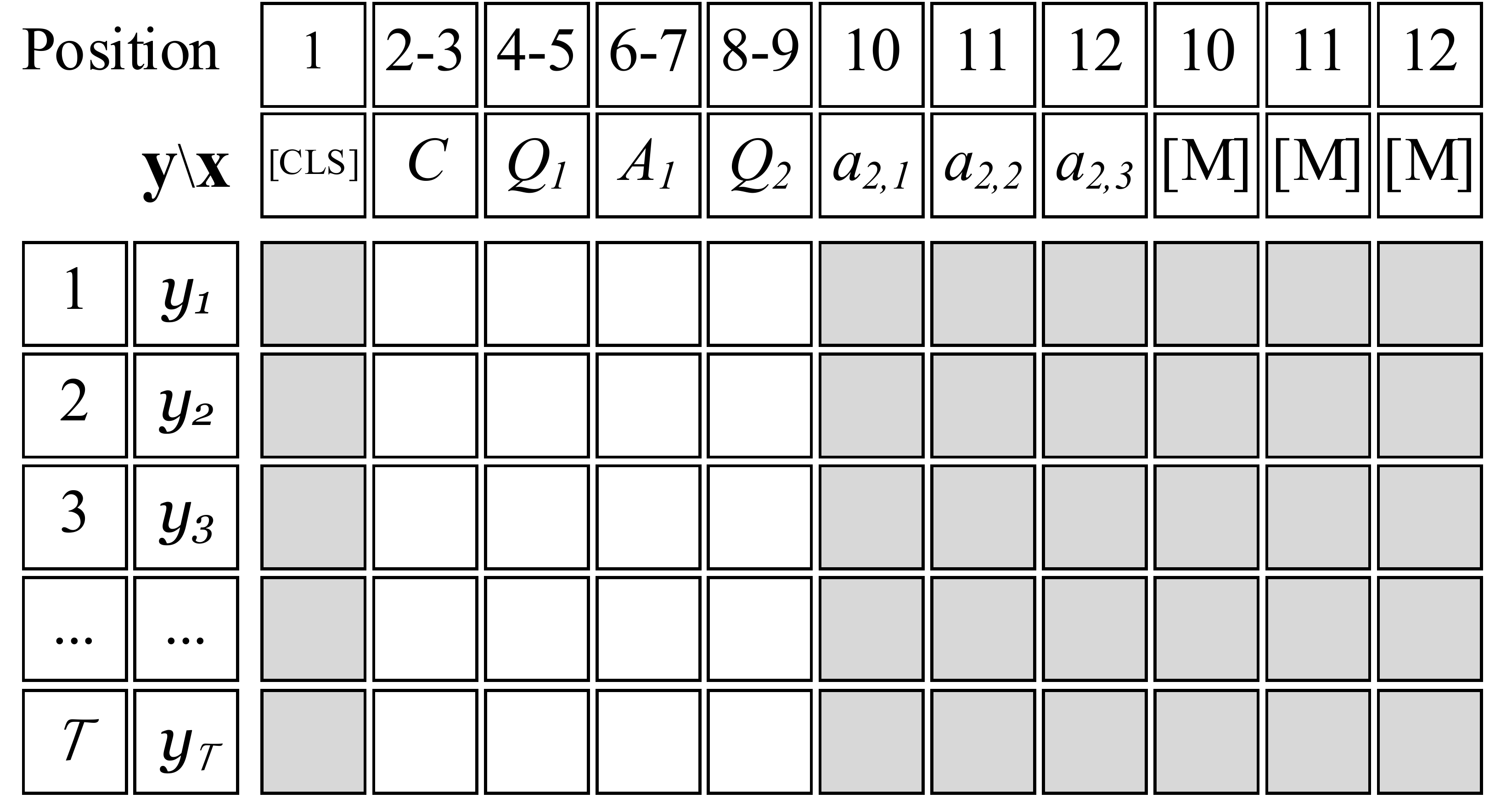}  
  \caption{Generative co-attention mask in visual stream.}  
  \label{fig3}
\end{figure} 
For the generative co-attention mask in visual stream with $t$=2 shown in Figure ~\ref{fig3}, to prevent the information of $\hat{A}_2$ from leaking through visual features during training, all visual tokens can access $H$=$\{C,Q_1,A_1\}$ and $Q_2$ but have no access to \texttt{[CLS]} and $\hat{A}_2$=$[a_{2,1},a_{2,2},a_{2,3}]$.
Besides, for the remaining two attention masks, to make linguistic tokens access all visual tokens, we set the elements of the co-attention mask matrix in the linguistic stream to all 0s. And to make all visual tokens access each other, we set the elements of the self-attention mask matrix in the visual stream to all 0s.

\subsection{Training Tasks in UniMM-UL}
\label{Language_Tasks} 
To enable the model to generate answers and take advantage of negative instances, the autoregressive MLM task with unlikelihood training is exploited in the generative setting.
Besides, autoencoding  MLM and NSP tasks are used in both settings for multi-task learning. Then, we perform the unified training with sparse annotations.
\subsubsection{Autoregressive Masked Language Modeling}
\label{AR_MLM}

The autoregressive MLM task is to predict the tokens of the answer in an autoregressive way. And the task is performed only in the generative setting to implement the answer generation. 

In addition to using likelihood training on positive instances, we utilize UnLikelihood (UL) training on negative ones.
Specifically, to learn useful knowledge from the negative instances and make up for the disadvantages of the likelihood objective, we take various negative instances as unlikely generations and use the unlikelihood training to force the model to assign a low probability to them. 

As the dialog example with $t$=2 shown in Figure~\ref{fig2} (c), based on the final representations of the last three \texttt{[M]} tokens at positions 10, 11, and 12, the autoregressive MLM task is to predict each token $[a_{2,1}, a_{2,2}, a_{2,3}]$ of $\hat{A}_2$. In general, for $\hat{A}_t \in \{A_t^{(k)}\}_{k=1}^{N_c}$, if $\hat{A}_t$ is correct, we use the likelihood objective to optimize the model. Otherwise, we use the unlikelihood objective to optimize the model to prevent the model from generating incorrect answers. Formally, the autoregressive MLM loss $\mathcal{L}_{\text{AR}}$ is obtained by:
\begin{equation}
\small
\label{L_AR}
\begin{split}
    \mathcal{L}_{\text{AR}} &= 
                \begin{cases}
                    \mathcal{L}_{\text{L}}(\hat{A}_t|I,H, Q_t), &\text{if $\hat{A}_t$ is correct}\\
                    \mathcal{L}_{\text{UL}}(\hat{A}_t|I,H, Q_t), &\text{otherwise}
                \end{cases}, \\
    \mathcal{L}_{\text{L}}(\hat{A}_t|I,H, Q_t) &= -\sum_{i=1}^{|\hat{A}_t|} \text{log }p(a_{t,i}|I,H, Q_t, a_{t,<i}),\\
    \mathcal{L}_{\text{UL}}(\hat{A}_t|I,H, Q_t) &= -\sum_{i=1}^{|\hat{A}_t|} \text{log } (1-p(a_{t,i}|I,H, Q_t, a_{t,<i})),
\end{split}
\end{equation}
where $a_{t,<i}$ is the generated sequence prior to $a_{t,i}$. $|\hat{A}_t|$ is the length of $\hat{A}_t$. $\mathcal{L}_{\text{L}}$ and $\mathcal{L}_{\text{UL}}$ are likelihood and unlikelihood objectives.

In the decoding process, we rely on the unified multimodal model with generative attention masks. Specifically, a mask \texttt{[M]} token is added at the end of the input to induce a one-step generation, and then the mask \texttt{[M]} token is replaced with the generated token for the next token generation. We use greedy search as the decoding strategy, terminating the decoding process when the \texttt{[SEP]} token is generated.
We use the log-likelihood scores of each token to rank all answer candidates.

\subsubsection{Autoencoding Masked Language Modeling}
\label{AE_MLM}
The autoencoding MLM task is to recover the masked tokens based on the surrounding language and image context and used as an auxiliary task to make the model understand the  context.
As shown in Figure~\ref{fig2} (c), we compute autoencoding MLM loss $\mathcal{L}_{AE}$ in both settings.   
In the discriminative setting, we randomly mask and recover 15\% of tokens in the whole dialog sequence \{$H$, $Q_t$, $A_t$\} and do not recover the masked tokens in $\hat{A}_t$ when $\hat{A}_t$ is incorrect to avoid noise caused by negative instances.
However, in the generative setting, since we perform the autoregressive MLM task on $\hat{A}_t$, we randomly mask and recover 15\% of the tokens only in $H$ and $Q_t$.

\subsubsection{Next Sentence Prediction} 
The NSP task is taken as a binary classification problem for answer discrimination in the discriminative setting and used as an auxiliary task in the generative setting.
The NSP loss $\mathcal{L}_{\text{NSP}}$ is computed in both settings in the same way. 
Specifically, the NSP head is over \texttt{[CLS]} and is trained to predict 1 when the input contains the correct answer and 0 when the input contains an incorrect answer sampled from candidate answers.

\subsubsection{Unified Training with Sparse Annotations}

\label{training_sparse}
The overall training objective is the sum of different types of objectives according to generative and discriminative settings.  
Specifically, within one training batch, both discriminative and generative settings are sampled at a scale of 1/2. 
In addition to the above language tasks, we also use a vision task, i.e., Masked Image Region Prediction (MIRP)~\cite{VisDialBert_2020_Vishvak}, and compute its loss $\mathcal{L}_{\text{MIRP}}$ by masking 15\% of the image features.
In the discriminative setting, we allow all tokens to attend to each other in each attention by setting mask matrices $\mathbf{M}$ to all 0s, and the training objective is the sum of $\mathcal{L}_{\text{AE}}$, $\mathcal{L}_{\text{NSP}}$, and $\mathcal{L}_{\text{MIRP}}$. 
In the generative setting, we use the proposed generative attention masks (Section~\ref{section_generative_masks}), and the training objective is the sum of $\mathcal{L}_{\text{AR}}$, $\mathcal{L}_{\text{AE}}$, $\mathcal{L}_{\text{NSP}}$ and $\mathcal{L}_{\text{MIRP}}$. 
Therefore, UniMM-UL could support two settings through the same model with shared parameters.

\subsection{Unified Fine-tuning with Dense Annotations}
\label{finetuning_dense}
For question $Q_t$, dense annotations provide dense relevance scores $\mathbf{S}=\{s(A_t^{(k)})\}_{k=1}^{N_c}$ for all candidate answers $\{A_t^{(k)}\}_{k=1}^{N_c}$. We adopt different fine-tuning methods in discriminative and generative settings to utilize dense annotations.

\subsubsection{Generative Dense Fine-tuning}
Our goal is to fine-tune the model to tend to generate candidates with positive relevance scores and reduce the probability of producing candidates whose relevance scores are 0s. Therefore, $\mathcal{L}_{\text{UL}}$ is computed using a candidate whose relevance score is 0, while $\mathcal{L}_{\text{L}}$ is computed using a candidate whose relevance score is greater than 0. In addition, we weight $\mathcal{L}_{\text{L}}$ according to the non-zero $s(A_t^{(k)})$.
Formally, we extend $\mathcal{L}_{\text{AR}}$ in Eq.~(\ref{L_AR}) to $\mathcal{L}_{\text{AR\_dense}}$ by:
\begin{equation}
\small
\begin{split} 
    \mathcal{L}_{\text{AR\_dense}} &= \sum_{k=1}^{N_c} \mathcal{L}_{\text{AR}}^{(k)}, \\
    \mathcal{L}_{\text{AR}}^{(k)} &= 
                \begin{cases}
                    s({A}_t^{(k)}) \cdot \mathcal{L}_{\text{L}}({A}_t^{(k)}|I,H, Q_t),  &\text{if $s({A}_t^{(k)})> 0$}\\
                    \mathcal{L}_{\text{UL}}({A}_t^{(k)}|I,H, Q_t),  &\text{otherwise}
                \end{cases}.
\end{split}
\end{equation}

\subsubsection{Discriminative Dense Fine-tuning}
The NSP head is used to to predict probabilities $\mathbf{P}=\{p(A_t^{(k)})\}_{k=1}^{N_c}$ for all candidate answers. 
According to previous work~\cite{VDBERT_2020_wang}, fine-tuning with dense annotations is regarded as a Learning To Rank (LTR) task.
Therefore, we select and compare several state-of-the-art LTR methods. 
Finally, we adopt neuralNDCG$^\mathrm{T}$~\cite{NeuralNDCG} with the best NDCG score as the ranking module for answering discrimination. 
Formally, we extend $\mathcal{L}_{\text{NSP}}$ to $\mathcal{L}_{\text{NSP\_dense}} =  \text{neuralNDCG}^{\mathrm{T}}(\mathbf{P}, \mathbf{S})$.

During fine-tuning, we replace $\mathcal{L}_{\text{NSP}}$ by $\mathcal{L}_{\text{NSP\_dense}}$, and replace $\mathcal{L}_{\text{AR}}$ by $\mathcal{L}_{\text{AR\_dense}}$. Since the difference between dense and sparse annotations only exists in language tasks, we do not mask and recover the image tokens for MIRP, i.e., $\mathcal{L}_{\text{MIRP}}=0$. The rest of the setup is the same as in section~\ref{training_sparse}.

\section{Experiments}
\subsection{Experiment Setup}
\subsubsection{Dataset and Evaluation Metric}
We conduct experiments on the VisDial v1.0 dataset~\cite{VisualDialog_2017_Das}. VisDial v1.0 contains about 130k human-human dialogs and is an extension of VisDial v0.9 with extra 10k dialogs.
Specifically, in VisDial v1.0, there are 123k/2k/8k dialogs of train/validation/test-standard set, and each dialog consists of one image, one caption, and $N_t$=10 question-answer pairs.
Each question is equipped with $N_c$=100 candidate answers, one of which is considered the correct answer in the sparse annotation.
For all rounds of all dialogs in all sets, VisDial v1.0 provides sparse annotations.   
Dense annotations of candidate answers are provided for only one round of a portion of the training set (2k dialogs), the entire validation set, and the entire test-standard set\footnote{
Since the annotations of the test-standard set are private, we could only do evaluation on the test-standard set on the online server with a maximum of five submissions per account.
We report discriminative results on the test-standard set and conduct other experiments (including generative evaluation, ablation study, and case study) that require annotations on the validation set.
}.

Following~\cite{VisualDialog_2017_Das}, the model is evaluated on retrieval metrics, which consist of Recall@1, Recall@5, Recall@10, Mean Reciprocal Rank (MRR), and Mean rank of the correct answer (Mean).  
Since obtaining dense annotations, the ranking metric Normalized Discounted Cumulative Return (NDCG) has become the primary metric to evaluate model performance in the VisDial challenge~\cite{VDBERT_2020_wang}.
The results on the test-standard set are displayed on the public leaderboard\footnote{\url{https://eval.ai/web/challenges/challenge-page/518/leaderboard/1421}}.

\subsubsection{Baselines} 

In the generatrive setting, we consider recent generative baselines, including KBGN~\cite{jiang2020kbgn}, DAM~\cite{jiang2020dam}, LTMI~\cite{nguyen2020LTMI}, MITVG~\cite{multimodal_2021_Chen}, GOG-Multi-Gen~\cite{gog_2021_chen}, and LTMI-LG~\cite{learningground_2021_chen}.
In the discriminative setting, we consider state-of-the-art discriminative baselines, including KBGN~\cite{jiang2020kbgn}, Modality-Balanced~\cite{kim2020modality}, FGA~\cite{schwartz2019fga}, DualVD~\cite{jiang2020dualvd}, MCA~\cite{agarwal2020historyneed}, P1+P2~\cite{Qi_2020_P1P2}, SGL~\cite{kang2021reasoningVD}, CARE~\cite{Exploring_2021_Li},
LTMI-GOG-Multi~\cite{gog_2021_chen},  VisDial-BERT~\cite{VisDialBert_2020_Vishvak}, VD-BERT~\cite{VDBERT_2020_wang}, and VD-PCR~\cite{VDPCR_2022_Yu}.

\subsubsection{Implementation details}

The model architecture of UniMM-UL follows that of VisDial-BERT~\cite{VisDialBert_2020_Vishvak} for a fair comparison. 
Specifically, for the visual stream, transformer layers number is $N_1$=6, attention heads number is 8, and hidden state size is 1024. 
For the linguistic stream, transformer layers number is $N_1$+$N_2$=$12$, attention heads number is 12, and hidden state size is 768. The co-attention layers connect the last $N_1$=6 layers in visual and linguistic streams.
In autoencoding MLM, we use the same mask probability and mask strategy as BERT~\cite{Devlin2019BERT, VDBERT_2020_wang}. 
In unified training, we utilize automatic mixed precision and distributed training to conduct all experiments on 4 GPUs (with a batch size of 240). 
And to fully exploit the number of negative instances, each positive one is equipped with 5 negative ones sampled from all 99 options in each epoch. 
Besides, during unified fine-tuning, due to GPU memory limitations, the model learns one sample containing 100 options each time, and gradient accumulation is used to make larger mini-batches (with a batch size of 16). The learning rate is $2\times10^{-5}$, with linear warm up over the first 2k steps and linear decay to $1\times10^{-5}$ over 200k steps. All models are implemented by PyTorch~\cite{Paszke2019pytorch}. Our code will be publicly available at: \url{https://github.com/ZihaoW123/UniMM}.

\subsection{Main Results}
 
\subsubsection{Generative Results} 
\begin{table}
  \caption{Generative results of ranking the candidates by the model’s  (a) \emph{sequence-level log-likelihood} scores and (b) \emph{token-level log-likelihood} scores on VisDial v1.0 validation set. $*$ denotes fine-tuning on dense annotations. "UniMM" denotes our model without Unlikelihood Training. Underline indicates the highest performance among prior models.} 
  \label{tab:gen_result1}
  \resizebox{0.92\linewidth}{!}{
  \begin{tabular}{l|l|cccccc}
    \toprule
    \multicolumn{2}{c|}{Model} & NDCG $\uparrow$  & MRR $\uparrow$ & R@1 $\uparrow$ & R@5 $\uparrow$ & R@10 $\uparrow$ & Mean $\downarrow$ \\
    \midrule   
    \multirow{6}{*}{(a)}   
    &KBGN~\cite{jiang2020kbgn} & 60.42 & 50.05 & 40.40 & 60.11 & 66.82 & 17.54  \\
    &DAM~\cite{jiang2020dam}  & 60.93 &50.51& 40.53& 60.84& 67.94& 16.65\\
    &LTMI~\cite{nguyen2020LTMI}  & \underline{63.58} & 50.74& 40.44& 61.61& 69.71&14.93\\
    &MITVG~\cite{multimodal_2021_Chen}  & 61.47& 51.14& 41.03& 61.25& 68.49&\underline{14.37}\\
    &GOG-Multi-Gen~\cite{gog_2021_chen} & 63.35& \underline{51.80}& \underline{41.78} & \underline{62.23}& 69.79&15.16\\
    &LTMI-LG~\cite{learningground_2021_chen} & 63.53 &51.43 & 41.68& 61.96& \underline{69.87}& 14.89 \\
    \cline{2-8}
    &UniMM & 63.89 & 52.29& 42.09& 62.96& 70.66&13.99\\
    &UniMM-UL & 62.86& \textbf{53.49}& \textbf{42.70}& \textbf{65.03}& \textbf{74.58}&\textbf{10.65}\\
    &UniMM-UL* &\textbf{69.23}& 51.31& 40.14& 63.16& 73.22& 11.53 \\
    \bottomrule 
    \toprule
    \multirow{3}{*}{(b)} &
    UniMM & 63.53 & 52.07 & 41.17 & 63.42 & 74.78 & 8.90 \\
    &UniMM-UL & 66.73& \textbf{57.52}& \textbf{44.23}& \textbf{72.84}& \textbf{85.80}&\textbf{5.21}\\
    &UniMM-UL* &\textbf{72.88}& 48.23& 32.53& 65.91& 81.46& 6.58 \\
    
   \bottomrule
\end{tabular}} 
\end{table} 

We rank the candidate answers using the log-likelihood scores and report the generative results on the VisDial v1.0 validation set in Table~\ref{tab:gen_result1}.

In previous work, the candidate answers are ranked by sequence-level log-likelihood scores. 
Formally, sequence-level log-likelihood score $L_{\text{seq}}$ of each candidate answer $\hat{A}_t$ is computed by: 
\begin{equation}
\label{LL}
\small
\begin{split}
    L_{\text{seq}} = \sum_{i=1}^{|\hat{A}_t|} \text{log }p(a_{t,i}|I,H, Q_t, a_{t,<i}), \text{ where } \hat{A}_t \in \{ A_t^{(k)}\}_{k=1}^{100}
\end{split}
\end{equation} 

As the more tokens in the answers, the more times of accumulation in the above formula, we find that the sequence-level log-likelihood scores of long answers are usually larger than that of short answers.
Therefore, to correct the influence of answer length, we additionally utilize the token-level log-likelihood score:
\begin{equation}
\label{LL}
\small
\begin{split}
    L_{\text{token}} = \frac{1}{|\hat{A}_t|}\sum_{i=1}^{|\hat{A}_t|} \text{log }p(a_{t,i}|I,H, Q_t, a_{t,<i}),
    \text{ where } \hat{A}_t \in \{ A_t^{(k)}\}_{k=1}^{100}
\end{split}
\end{equation}

By making the comparison, we make the following observations.

\paragraph{New state-of-the-art generative results}
Our models achieve the best scores on metrics among all published generative results. 
Specifically, UniMM, which does not use unlikelihood training, outperforms previous work on all metrics, showing the superiority of the unified model with multi-task learning. 
In addition, since VD-BERT does not report generative results on VisDial V1.0, we adapt GOG-Multi-Gen~\cite{gog_2021_chen} and MITVG~\cite{multimodal_2021_Chen} as baselines that outperform VD-BERT.
With unlikelihood training, UniMM-UL achieves the best scores on retrieval metrics, indicating UniMM-UL tends to generate correct answers rather than incorrect answers.  
After fine-tuning with dense annotations, UNIMM-UL$^*$ obtains the best score (69.23) on the ranking metric NDCG, indicating the model successfully generates relevant answers by exploiting dense annotations.

\paragraph{Unlikelihood training and generative fine-tuning method are both effective}
In Table~\ref{tab:gen_result1} (a) and (b), UniMM-UL outperforms UniMM on all retrieval metrics, indicating unlikelihood training helps the model reduce the probability of generating incorrect answers. Besides, NDCG scores of UniMM-UL$^*$ are higher than that of UniMM and UniMM-UL, indicating our generative fine-tuning method is beneficial for the model to generate highly relevant answers.   

\paragraph{Generative results are sensitive to token-level log-likelihood scores.} 
The gaps among UniMM, UniMM-UL, and UniMM-UL* using $L_{\text{token}}$ are larger than that using $L_{\text{seq}}$. For example, in Table~\ref{tab:gen_result1} (a), the difference in MMR score between UniMM and UniMM-UL is 1.20, but in Table~\ref{tab:gen_result1} (b), the difference increases to 5.45.
Therefore, to analyze the impact of each module on the model, $L_{\text{token}}$ can be used to widen the distance among various ablation models.

\subsubsection{Discriminative Results}
\begin{table}
  \caption{Discriminative results on VisDial v1.0 test-standard set. ${\dag}$ denotes ensemble model. $*$ denotes fine-tuning on dense annotations. $\triangle$ denotes that the model utilizes extra pronoun coreference resolution labels for training. 
  Underline denotes the highest performance among prior methods.
  }
  \label{tab:dis_result1}
  \resizebox{0.92\linewidth}{!}{
  \begin{tabular}{lcccccc}
    \toprule
    Model & NDCG $\uparrow$  & MRR $\uparrow$ & R@1 $\uparrow$ & R@5 $\uparrow$ & R@10 $\uparrow$ & Mean $\downarrow$ \\
    \midrule  
    
    KBGN~\cite{jiang2020kbgn} &  57.60 & 64.13 & 50.47 & 80.70 & 90.16 & 4.08 \\
    FGA$^{\dag}$~\cite{schwartz2019fga} & 54.50 & 67.30 & 53.40 & 85.28 & 92.70 & 3.54  \\
    Modality-Balanced$^{\dag}$~\cite{kim2020modality} &59.49 &64.40 &50.90 &81.18 &90.40& 3.99\\ 
    DualVD~\cite{jiang2020dualvd} &56.32&63.23 &49.25 &80.23 & 89.70&4.11 \\
    SGL~\cite{kang2021reasoningVD} & 61.97 & 62.28 & 48.15 & 79.65 & 89.10 & 4.34\\
    LTMI-GOG-Multi~\cite{gog_2021_chen} & 61.04& 63.52& 50.01& 80.13& 89.28&4.31\\ 
    CARE~\cite{Exploring_2021_Li} & 60.60& 64.78& 51.60& 80.70& 90.60 &4.07 \\  
    VD-BERT~\cite{VDBERT_2020_wang} &59.96& 65.44& 51.63& 82.23& 90.68& 3.90\\ 
    VD-PCR$^\triangle$~\cite{VDPCR_2022_Yu}& 63.55 & \underline{68.73} & \underline{55.45}& \underline{85.38}& \underline{93.53}& \underline{3.21} \\  
    \midrule  
    
    VisDial-BERT~\cite{VisDialBert_2020_Vishvak} & \underline{63.87}& 67.50& 53.85& 84.68& 93.25&3.32\\ 
    UniMM-UL & \textbf{63.90}& 68.14& 54.57& 85.15& 93.13&3.27\\
    
    \bottomrule 
    \toprule
    MCA$^{*}$~\cite{agarwal2020historyneed}& 72.47& 37.68& 20.67& 56.67& 72.12&8.89\\ 
    P1+P2$^{*}$~\cite{Qi_2020_P1P2}& 71.60 & 48.58 & 35.98 & 62.08 & 77.23 & 7.48\\  
    VD-BERT$^{*}$~\cite{VDBERT_2020_wang}& 74.54& 46.72& 33.15& 61.58& 77.15&7.18\\
    VD-PCR$^{\triangle*}$~\cite{VDPCR_2022_Yu} &\underline{75.30}& \underline{56.17}& \underline{45.32}& \underline{68.10}&\underline{82.30} &\underline{5.95} \\

    \midrule 
    VisDial-BERT$^*$~\cite{VisDialBert_2020_Vishvak} &74.47&50.74 & 37.95& 64.13& 80.00& 6.28\\
    UniMM-UL$^{*}$ & \textbf{75.92}& \textbf{56.18}& 43.70& \textbf{71.03}& \textbf{84.80}&\textbf{5.42}\\ 
    
   \bottomrule
   \toprule

    P1+P2$^{*\dag}$~\cite{Qi_2020_P1P2}& 74.02 & 52.62 & 40.03 & 68.85 & 79.15 & 6.76\\
    
    VD-BERT$^{*\dag}$~\cite{VDBERT_2020_wang}& 75.35& 51.17& 38.90& 62.82& 77.98& 6.69\\ 
    VD-PCR$^{\triangle*\dag}$~\cite{VDPCR_2022_Yu} &\underline{76.14}& \underline{56.05}& \underline{44.75}& \underline{68.40}&\underline{82.75} &\underline{5.72} \\
    \midrule 
    UniMM-UL$^{*\dag}$ & \textbf{76.17}& \textbf{56.42}& 44.32& \textbf{70.23}& \textbf{84.52}&\textbf{5.47}\\
   \bottomrule
\end{tabular}}
\end{table}

We further report the discriminative results on the test-standard set of VisDial v1.0 in Table~\ref{tab:dis_result1} and make the following findings and discussions.

\paragraph{Comparable discriminative results in both single-model and ensemble settings.}
For the models trained with sparse annotations, the goal is to retrieve the only correct answer among all candidate answers. Compared to the VisDial-BERT baseline, our unified model UniMM-UL gains much improvement on all metrics. And UniMM-UL also outperforms most published baselines except VD-PCR$^\triangle$, which utilizes additional PCR labels for training. 
For the models fine-tuned with dense annotations, the target is to maximize the ranking metric NDCG.   
In the single-model setting, compared with all baselines, UniMM-UL$^*$ obtains a better NDCG of 75.90. 
Besides, we report an ensemble version UniMM-UL$^{*\dag}$ of 5 models, which differ only in the initial seed, which also achieves the best NDCG of 76.17 among all published results.
The results indicate the success of our proposed model in the discriminative setting.  

\paragraph{Our UniMM-UL is more effective and powerful than VisDial-BERT}

As VisDial-BERT also adopts ViLERT as the base model, the UniMM-UL model without generative setting degenerates into the original VisDial-BERT that only supports answer discrimination. 
As shown in Table~\ref{tab:dis_result1}, UNIMM-UL is superior to VisDial-BERT on all metrics, and after fine-tuning with dense annotations, UniMM-UL$^*$ still outperforms VisDial-BERT$^*$ on all metrics.
The improvements indicate the effectiveness of incorporating discriminative and generative settings via unified training and the compatibility between the tasks. 
Moreover, the generative and discriminative results demonstrate the effectiveness of our UniMM-UL in both generative and discriminative settings using a two-stream Transformer. By contrast, VisDial-BERT and VD-PCR only support the discriminative setting.


\paragraph{Advances in our fine-tuning methods with dense annotations.}

When training the model with only sparse annotations, the NDCG of UniMM-UL (63.90) is slightly higher than that of VisDial-BERT (63.87) by 0.03. However, after fine-tuning with dense annotations, the gap on NDCG between UniMM-UL$^*$ (75.92) and VisDial-BERT$^*$ (74.47) increases to 1.45. Besides, though VD-PCR$^\triangle$ using extra PCR labels performs better UniMM-UL on retrieval metrics, after fine-tuning with dense annotations, our UniMM-UL$^*$ achieves the best results on all ranking and retrieval metrics except R@1. Especially, our UniMM-UL$^*$ and UniMM-UL$^{*\dag}$ sets new state-of-the-art NDCG records of 75.92 and 76.17 in single-model and ensemble settings. The significant improvement on the NDCG metric demonstrates the effectiveness of our proposed fine-tuning methods.

\paragraph{Why the proposed method cannot achieve a promising result in some retrieval metrics in some cases}
On the metric R@1, UniMM-UL$^*$ and UniMM-UL$^*\dag$ perform worse than VD-PCR$^{\triangle*}$ and VD-PCR$^{\triangle*\dag}$, respectively. One important reason is that VD-PCR adds the PCR task using extra visual dialog PCR (pronoun coreference resolution) labels. 
And VD-PCR is jointly trained with coreference-related attention head selection. 
If a pronoun appears in the question, PCR labels help the model to associate the pronoun with its corresponding noun. 
Therefore, VD-PCR has an advantage in selecting the best answer if there are pronouns in the question, and performs better than our model on some retrieval metrics. 

\subsection{Ablation Study} 
\begin{table}
  \caption{Ablation studies in the discriminative setting on val v1.0: dense annotation fine-tuning with (a) various settings and (b) various ranking methods.} 
  \label{tab:disc_dense_ablation}
  \resizebox{0.92\linewidth}{!}{
  \begin{tabular}{l|l|cccccc}
    \toprule
    \multicolumn{2}{c|}{Model} & NDCG $\uparrow$  & MRR $\uparrow$ & R@1 $\uparrow$ & R@5 $\uparrow$ & R@10 $\uparrow$ & Mean $\downarrow$ \\
    \midrule    
    \multirow{3}{*}{(a)}   
    &UniMM-UL$^*$ & \textbf{77.60} & \textbf{58.68} & \textbf{46.81}  & 72.83 & 85.16 & 5.17  \\ 
    &w/o gen. setting & 77.45 & 57.49 & 45.95 & 70.64 & 83.59 & 5.45 \\   
    &w/o $\mathcal{L}_{\text{UL}}$ & 77.29 & 58.10 & 45.58 & \textbf{72.84} & \textbf{85.81} & \textbf{5.05} \\ 
    \bottomrule 
    \toprule
    \multirow{6}{*}{(b)} &
    CE & 76.97 & 55.41 & 42.95 & 69.21 & 82.84 & 5.62 \\
    &ListNet & 76.70 & 55.96 & 42.97 & 70.87 & 84.25 & 5.39 \\
    &ListMLE & 76.46 & 59.27 & 47.76 & 72.11 & 84.99 & 5.12  \\
    &ApproxNDCG & 75.71 & 57.99 & 42.64 & \textbf{75.90}  & \textbf{86.71} & \textbf{4.78}  \\
    &NeuralNDCG & 77.34 & \textbf{59.51} & \textbf{47.87} & 73.24 & 85.21 & 5.11 \\
    &NeuralNDCG$^\mathrm{T}$ & \textbf{77.60} & 58.68 & 46.81  & 72.83 & 85.16 & 5.17 \\
    
   \bottomrule
\end{tabular}} 
\end{table}

\begin{table} 
  \caption{Ablation studies in the generative setting on val v1.0: dense annotation fine-tuning with various settings. The candidates are ranked by the model’s (a) \emph{sequence-level log-likelihood} scores and (b) \emph{token-level log-likelihood} scores}.   
  \label{tab:gen_dense_ablation}
  \resizebox{0.92\linewidth}{!}{
  \begin{tabular}{l|ll|cccccc}
    \toprule
      \multicolumn{3}{c|}{Model} & NDCG $\uparrow$  & MRR $\uparrow$ & R@1 $\uparrow$ & R@5 $\uparrow$ & R@10 $\uparrow$ & Mean $\downarrow$ \\
    \midrule  
      \multirow{4}{*}{(a)} 
      &\texttt{[1]} & UniMM-UL$^*$ &\textbf{69.23}& \textbf{51.31}& \textbf{40.14}& \textbf{63.16}& \textbf{73.22}& \textbf{11.53} \\
      &\texttt{[2]} & \texttt{[1]-}disc. setting & 68.71 & 49.65  & 37.94 & 62.31 & 72.40 & 12.32 \\
      &\texttt{[3]} & \texttt{[2]-}$\mathcal{L}_{\text{NSP\_dense}}$  & 66.47 & 49.78  &  37.82 &  62.69 & 72.29  & 12.36 \\ 
      &\texttt{[4]} & \texttt{[3]-}$\mathcal{L}_{\text{UL}}$ & 67.61 & 48.87 & 38.16 & 59.95 & 68.15 & 13.49 \\ 
    \bottomrule
    \toprule 
      \multirow{4}{*}{(b)} 
      &\texttt{[1]} & UniMM-UL$^*$ &\textbf72.88& \textbf{48.23}& \textbf{32.53}& \textbf{65.91}& \textbf{81.46}& \textbf{6.58}   \\
      &\texttt{[2]} & \texttt{[1]-}disc. setting & \textbf{73.55} &  44.77 & 27.74  &  64.60 & 80.68  & 6.87 \\
      &\texttt{[3]} & \texttt{[2]-}$\mathcal{L}_{\text{NSP\_dense}}$  & 72.21 &  43.83 &  26.03 &  65.37 &  81.37 & 6.71 \\ 
      &\texttt{[4]} & \texttt{[3]-}$\mathcal{L}_{\text{UL}}$ & 65.95 & 39.51 & 25.06 & 54.70 & 69.65 & 9.63 \\ 
  \bottomrule
\end{tabular}}    
\end{table}

\begin{figure*}[t]
    \centering   
    \subfloat[]{\includegraphics[width=0.40\linewidth]{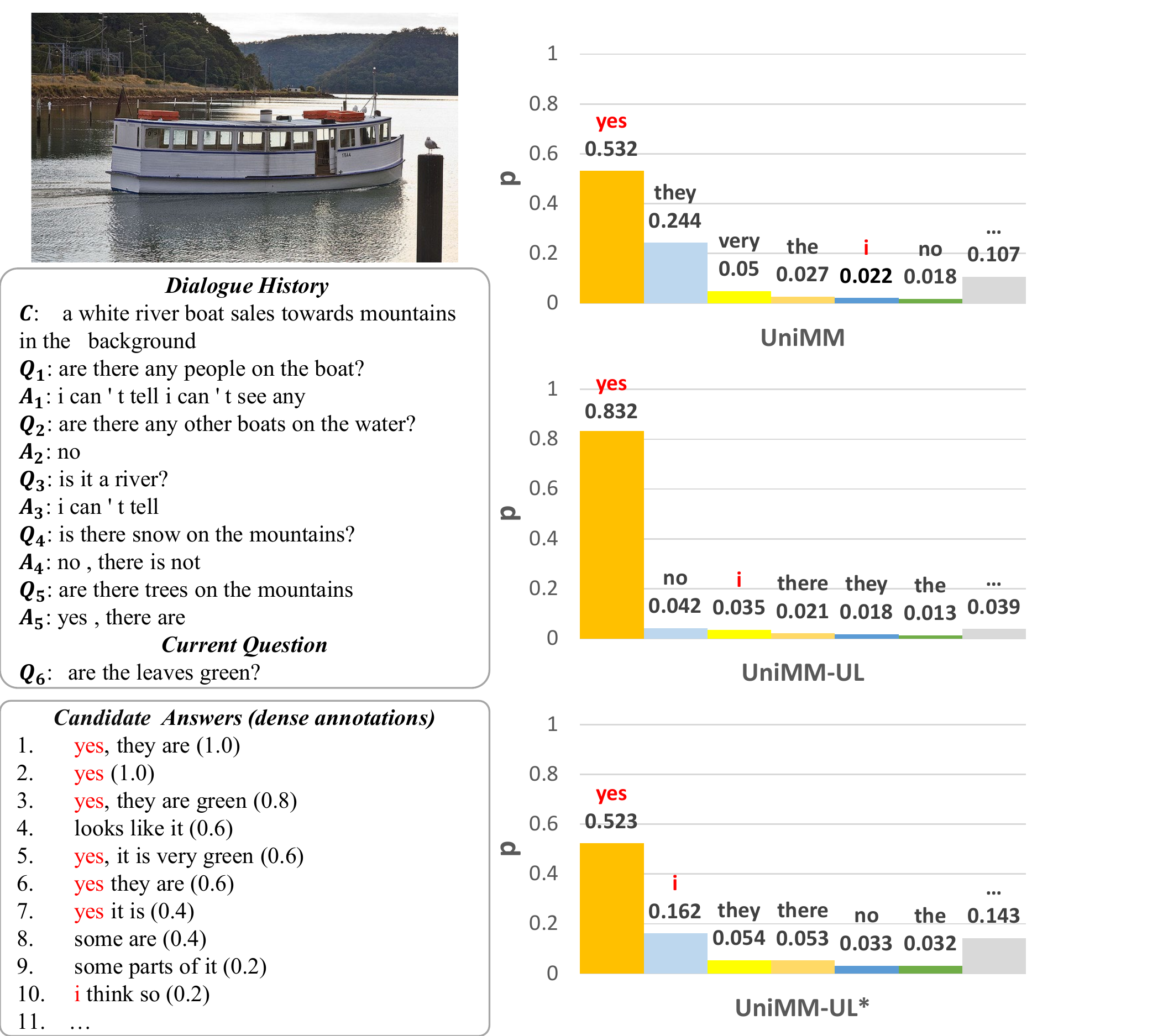}} 
    \subfloat[]{\includegraphics[width=0.40\linewidth]{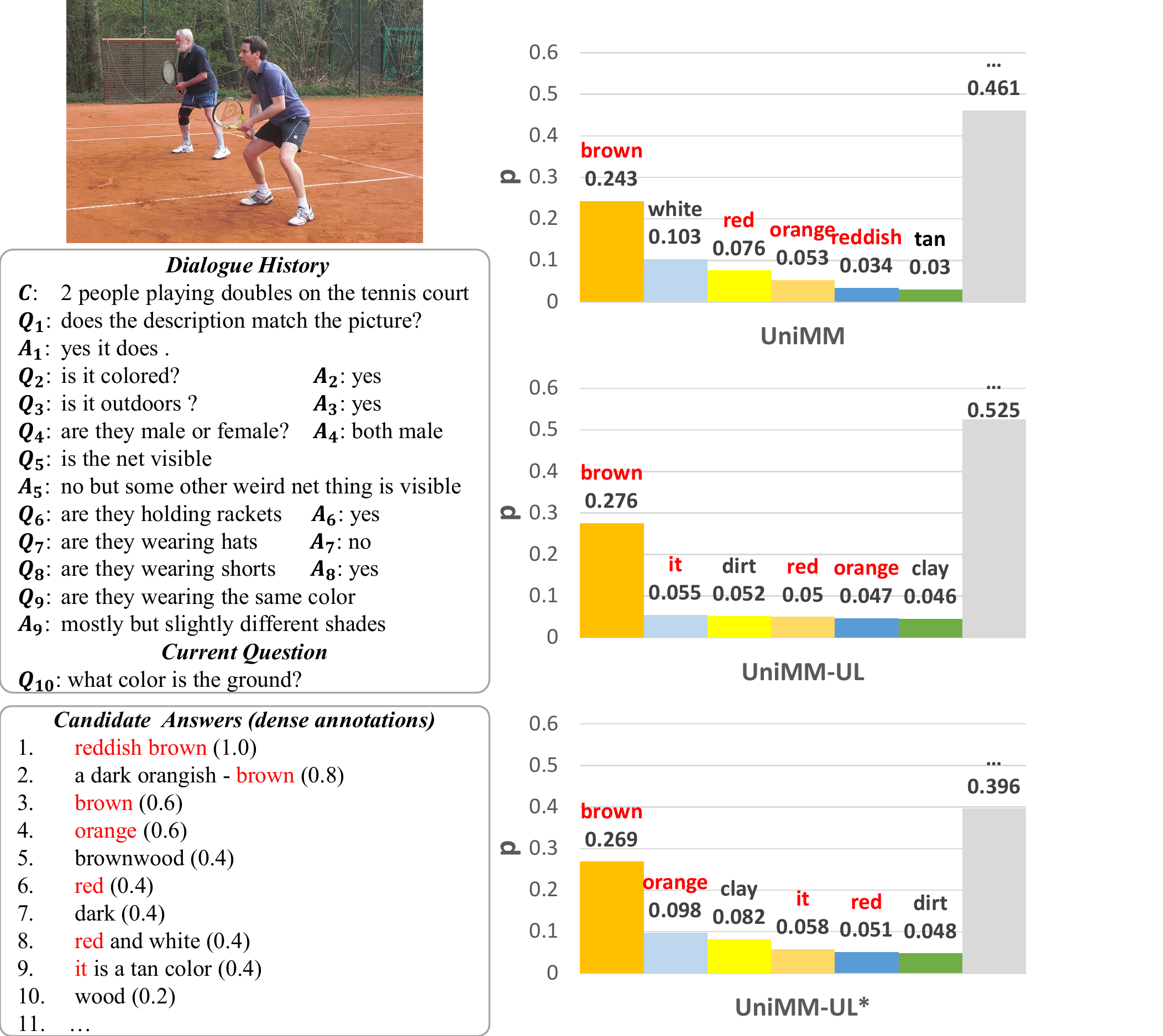}}
    \caption{Two examples from the VisDial v1.0 validation set are shown to compare UniMM, UniMM-UL, and UniMM-UL$^*$ in the generative setting. The probability distributions of the first token generated by various models are shown. In each probability histogram, we label the top 6 tokens with their probabilities, and "..." denotes the sum of the remaining tokens' probabilities. Key words occurring simultaneously in the probability histogram and dense annotations are highlighted in red.}  
    \label{fig:cases}
\end{figure*}

In this section, we study how different fine-tuning settings affect the discriminative and generative results.

\subsubsection{Ablation study on discriminative results}
We first study how different fine-tuning settings and ranking methods affect the discriminative results in Table~\ref{tab:disc_dense_ablation}(a).  
After removing the generative setting, the discriminative results are all worse, which shows that unified fine-tuning in both settings is more effective than only in the discriminative setting.
After only removing unlikelihood training in the generative setting, while R@5, R@10, and Mean become better, NDCG and MRR become worse, suggesting that the effect of likelihood training on discriminative results is not clear. 
We select a series of LTR modules as fine-tuning methods: Cross-Entropy (CE), ListNet~\cite{ListNet}, ListMLE~\cite{ListMLE}, approxNDCG~\cite{approxNDCG}, nertalNDCG, and nertalNDCG$^\mathrm{T}$~\cite{NeuralNDCG}. Among these methods, nertalNDCG$^\mathrm{T}$ yields the best NDCG. Therefore, we adopt the approxNDCG$^\mathrm{T}$ as our discriminative fine-tuning method.

\subsubsection{Ablation study on generative results}
We study how different fine-tuning settings affect the generative results in Table~\ref{tab:gen_dense_ablation}.  
After removing the discriminative setting, the generative results get worse on most metrics, which indicates that unified fine-tuning in both settings is also more effective than only in the generative setting.
After further removing $\mathcal{L}_{\text{NSP\_dense}}$ and $\mathcal{L}_{\text{UL}}$, the results become much worse on most metrics, which indicates that multi-task learning and unlikelihood training are beneficial to the fine-tuned model.

\subsection{Case Study}

In Figure~\ref{fig:cases}, we show two examples to qualitatively indicate how unlikelihood training and dense fine-tuning generate a good answer. In each histogram, the sum of all tokens’ probabilities is 1.0.
In example (a), compared with UniMM, UniMM-UL assigns a higher probability to the “yes” token and lower probabilities to others, suggesting that likelihood training helps the model increase the probabilities of correct answer tokens and decrease the probabilities of other tokens. 
In example (b), the "white" token's probability, which is the second-highest probability in UniMM, is reduced in both UniMM-UL and UniMM-UL$^*$, indicating that likelihood training reduces the probabilities of wrong tokens.
Besides, compared with UniMM, UniMM-UL$^*$ assigns higher probabilities to the "brown" and "orange" tokens that appear in candidate answers, indicating that fine-tuning with dense annotations makes the model assign high probabilities to relevant answers.  

\section{Conclusion} 

In this paper, we design a unified model UniMM-UL, which exploits both positive and negative instances for visual dialog understanding and generation. 
With the help of our unified training and fine-tuning methods, our unified model achieves state-of-the-art experimental results. Through ablation study and case study, we find that multi-task learning, unlikelihood training, and our fine-tuning methods are beneficial to the visual dialog model.  
 
Furthermore, since the pre-trained faster R-CNN is frozen during training, our model is not trained in an end-to-end manner, resulting in the limition of the capacity of the model and the time-consuming inference process. In the future, we will exploit the performance of our method in those end-to-end models~\cite{xu2021e2e, Dou_2022_CVPR} that achieve better performance while maintaining fast inference speed.

\begin{acks}
This work was supported by the National Key Research and Development Program of China (Grant No. 2018YFB2100801).
\end{acks}

\bibliographystyle{ACM-Reference-Format}
\bibliography{mybib}


\appendix

\section{Comparison between gen models}
 
\begin{figure}[b]
  \centering
  \includegraphics[width=1.0\linewidth]{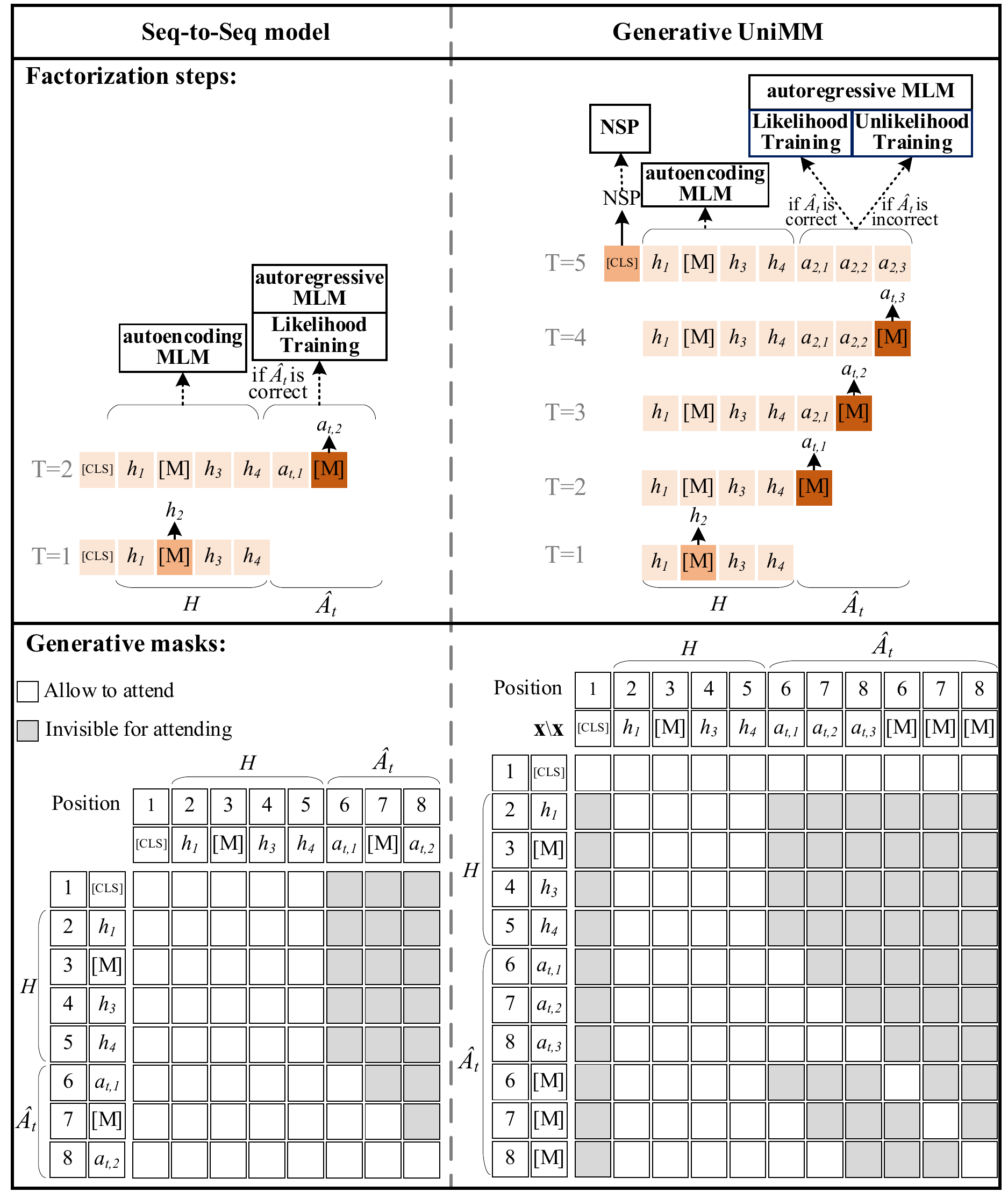}  
  \caption{
     Comparisons between generative UniMM-UL and Seq-to-Seq model (input sequence contains only language tokens). In the example $H = h_1, h_2, h_3, h_4$ and $\hat{A_t} = a_{t,1},  a_{t,2}, a_{t,3}$, the token $h_2$ of $H$ is masked by the special tokens \texttt{[M]}. For Seq-to-Seq model, the token $a_{t,2}$ of $\hat{A_t}$ is masked by \texttt{[M]}. For UniMM-UL, we show the generative self-attention mask in linguistic stream. To make a intuitive comparison, we ignore visual tokens and keep only text tokens.
 }  
\label{comparison}
\end{figure}

\label{appendix}
 
To further illustrate the contribution and novelty of our proposed generative attention mask. Here, we take the VD-BERT~\cite{VDBERT_2020_wang} as an example to make a detailed comparison between our model and the existing attention mechanism in the generative setting. To autoregressively generate an answer, VD-BERT uses the Seq-to-Seq attention mask proposed by UniLM~\cite{Li2020unilmv1}.
For each model, as shown in Figure~\ref{comparison}, we construct a new cloze instance for all factorization steps (top) and show the attention masks (bottom). 

Compared to the Seq-to-Seq self-attention mask used by VD-BERT, the technical differences of our generative attention mask are listed as follows:
First, in the generative setting, our model supports the NSP task, but VD-BERT can not. In our generative attention masks, ~\texttt{[CLS]]} can attend to all tokens, but other tokens can not attend to ~\texttt{[CLS]]}. By contrast, Seq-to-Seq model does not use ~\texttt{[CLS]]} token in the generative setting. Second, to implement the autoregressive MLM task, our model learns to autoregressively predict all (100\%) tokens of $\hat{A_t}$ (i.e., $a_{t,1}$,  $a_{t,2}$, $a_{t,3}$) during training, but Seq-to-Seq model only learns 15\% tokens of $\hat{A_t}$ (i.e., only $a_{t,2}$). Third, our model is a two-stream transformer, and VD-BERT and UniLM are single-stream transformers. We design different attention masks for two streams.







\end{document}